\def\year{2020}\relax
\documentclass[letterpaper]{article} 
\usepackage{aaai20}  
\usepackage{times}  
\usepackage{helvet} 
\usepackage{courier}  
\usepackage[hyphens]{url}  
\usepackage{graphicx} 
\usepackage{graphicx}  
\usepackage{booktabs}
\usepackage{multirow}
\usepackage{multicol}
\title{Towards Minimal Supervision BERT-based Grammar Error Correction}
\author{Yiyuan Li, Antonios Anastasopoulos, and Alan W Black\\
Carnegie Mellon University\\ 
5000 Forbes Avenue\\
Pittsburgh, Pennsylvania 15213\\
yiyuanli@andrew.cmu.edu \{aanastas, awb\}@cs.cmu.edu 
}
 
\begin{document}

\maketitle

\begin{abstract}
Current grammatical error correction (GEC) models typically consider the task as sequence generation, which requires large amounts of annotated data and limit the applications in data-limited settings. We try to incorporate contextual information from pre-trained language model to leverage annotation and benefit multilingual scenarios. Results show strong potential of Bidirectional Encoder Representations from Transformers (BERT) in grammatical error correction task.
\end{abstract}

\section{Introduction}

The goal of grammatical error correction is to detect and correct all errors in the sentence and return the corrected sentences. Current grammatical error correction approaches require a large amount of training data to achieve reasonable results, which unfortunately cannot be extended to languages with limited data. Recently, unsupervised models pre-trained on large corpora have boosted performance in many natural language processing tasks, which indicates a potential of leveraging such models for GEC in any language. In this work, we try to utilize (multilingual) BERT~\cite{devlin-etal-2019-bert} in order to perform grammatical error correction with minimum supervision.

\section{Proposed Method}
Our approach divides the GEC task into two stages: error identification followed by correction.
In the first stage, we try to detect the span in the original text that the edit will apply. We formulate this as a sequence labelling task where tokens are labelled in one of the following labels \{\textit{remain, substitution, insert, delete}\}. 
For the second stage (correction) we employ a pretrained-model like BERT. The labels from the error identification stage guide the masking of the inputs (where we mask any tokens marked with \textit{substitution} or insert mask tokens for \textit{insert} labels), and we obtain candidate outputs for every masked token. 
A shortcoming of our current approach is that it only produces as many corrections as masked input tokens; however, most grammar errors in fact tend to be edits of length~1 or~2, which are captured by our identification labels. The second stage is addressed in future work.

\section{Results}
We report preliminary results on the test part of the English FCE dataset~\cite{yannakoudakis-etal-2011-new}. The edits are labelled and scored through ERRANT~\cite{bryant-etal-2017-automatic}.

For our preliminary experiments we focus on a simplified single-edit setting, where we attempt to correct sentences with a single error (assuming oracle error identification annotations). The goal is to assess the capabilities of pre-trained BERT-like models assuming perfect performance for all other components (e.g. error identification).

We expand the original dataset to fit our single-error scenario with the following two schemes. (1) \textbf{each edit}: all corrections except one are applied, creating a single-error sentence; (2) \textbf{last edit}: all corrections except for the last one are applied. After the split, we obtain 3,585 and 1,024 corrections respectively. We also employ different strategies for deciding the number of masked tokens: (1) based on span length of the original edit, (2) based on length of the final correction (given from an oracle), and (3) using a single mask and measure whether any token of the correction is predicted. Note that subword predictions like \{ad, \#\#e, \#\#quate\} to adequate are merged in sentence-level, but remains in mask-level evaluation. 
Our preliminary results under the various settings are outlined in Table~\ref{tab: span-based evaluation single edit}. We find that different masking strategies have comparable performance, with slightly higher accuracy when using a single mask. Interestingly, BERT seems capable to actually produce corrections with quite high precision of more than~70\%.

\begin{table}[ht]
    \small
    \centering
    \scalebox{0.85}{
    \begin{tabular}{c|ccc|ccc}
    \toprule
        \multirow{2}{*}{Masking Strategy} & \multicolumn{3}{c|}{each edit} &
        \multicolumn{3}{c}{last edit}\\
        \cmidrule{2-7}
        & P@1 & R@1 & F$_{0.5}$@1 & P@1 & R@1 & F$_{0.5}$@1\\
        \midrule
        \# origin & 0.632 & 0.853 & 0.667 & 0.592 & 0.824 & 0.627\\
        \midrule
        \# target & 0.66 & 0.887 & 0.696 & 0.614 & 0.855 & 0.651\\
        \midrule
        single & \textbf{0.763} & \textbf{0.931} & \textbf{0.790} & \textbf{0.767} & \textbf{0.920} & \textbf{0.794}\\
        \bottomrule
    \end{tabular}}
    \caption{Sentence-level evaluation with different masking strategies for single edit pairs. Subword predictions are merged in sentence generation.}
    \label{tab: span-based evaluation single edit}
\end{table}
\begin{table}[ht]
    \small
    \centering
    \scalebox{0.85}{
    \begin{tabular}{c|cc|cc}
    \toprule
        \multirow{2}{*}{Masking Strategy} & \multicolumn{2}{c|}{each edit} &
        \multicolumn{2}{c}{last edit}\\
        \cmidrule{2-5}
        & Acc@1 & Acc@5 & Acc@1 & Acc@5\\
        \midrule
        \# origin & 0.292 & 0.455 & 0.229 & 0.390\\
        \midrule
        \# target & 0.313 & 0.484 & 0.247 & 0.405\\
        \midrule
        single & \textbf{0.365} & \textbf{0.554} & \textbf{0.312} & \textbf{0.501}\\
        \bottomrule
    \end{tabular}}
    \caption{A reranking mechanism could lead to better results, as performance@5 is higher than performance@1.}
    \label{tab: mask-based evaluation single edit}
\end{table}

We also study whether the correct output is among the top-$k$ candidates suggested by BERT. We compare the $F_{0.5}$ metrics based on each mask between the most probable prediction and the top k candidates, where k is set to 5. From the result in Table~\ref{tab: mask-based evaluation single edit}, we observe that an appropriate reranking model could further boost the performance by selecting the appropriate correction. 

\section{Related and Future work}
\begin{table}[ht]
    \small
    \centering
    \scalebox{0.9}{
    \begin{tabular}{@{}lp{0.45\textwidth}@{}}
       \toprule
       \multicolumn{2}{l}{\textbf{Example \#1: Redundant Edits}}\\
       \midrule
        \textbf{Source} & Of course there 's also a number 8 bus in front of the hotel , which is also suitable , but it leaves only every half an hour\\
        \textbf{Mask.} & Of course there 's also a number 8 bus [MASK] in front of the hotel , which is also suitable , but it leaves only every half an hour\\
        \textbf{Target} & Of course there 's also a number 8 bus \underline{,} in front of the hotel , which is also suitable , but it leaves only every half an hour\\
        \textbf{Ours} &  Of course there 's also a number 8 bus \underline{stop} in front of the hotel , which is also suitable , but it leaves only every half an hour\\
        \midrule
       \multicolumn{2}{l}{\textbf{Example \#2 : Synonyms}}\\
       \midrule
        \textbf{Source} & The aim of this report is to \underline{recomend} you to visit the Fuerte de San Diego Museum \\
        \textbf{Mask.} & The aim of this report is to [MASK] you to visit the Fuerte de San Diego Museum \\
        \textbf{Target} & The aim of this report is to \underline{recommend} you to visit the Fuerte de San Diego Museum \\
        \textbf{Ours} & The aim of this report is to \underline{allow} you to visit the Fuerte de San Diego Museum \\
        \midrule
        \multicolumn{2}{l}{\textbf{Example \#3 : Hallucination}}\\
        \midrule
        \textbf{Source} & Of course there 's also a \underline{bus number 8} , in front of the hotel , which is also suitable , but it leaves only every half an hour\\
        \textbf{Mask.} &  Of course there 's also a [MASK] [MASK] [MASK], in front of the hotel , which is also suitable , but it leaves only every half an hour\\
        \textbf{Target} & Of course there 's also a \underline{number 8 bus} , in front of the hotel , which is also suitable , but it leaves only every half an hour\\
        \textbf{Ours} & Of course there 's also a \underline{small parking station} , in front of the hotel , which is also suitable , but it leaves only every half an hour\\
    \bottomrule
    \end{tabular}}
    \caption{Common BERT prediction errors (we \underline{highlight} the original error and the prediction).}
    \label{tab: edit-error-analysis}
\end{table}

A retrieve-edit model is proposed for text generation \cite{Guu2017GeneratingSB}. However, the edition is one-time and sentences with multiple grammatical errors could further reduce the similarity between the correct form and the oracle sentence. An iterative decoding approach \cite{ge-etal-2018-fluency} or the neural language model \cite{choe-etal-2019-neural} as the scoring function are employed for GEC. To the best of our knowledge, there is no prior work in applying pre-trained contextual model in grammatical error correction. In the future work, we will additionally model \textit{error fertility}, allowing us to exactly predict the number of necessary [MASK] tokens. Last, we will employ a re-ranking mechanism which scores the candidate outputs from BERT, taking into account larger context and specific grammar properties.

\paragraph{Better span detection} Although BERT could predict all the missing token in the sentence in a reasonable way, prediction of the correct words could easily fall into redundant editing. Our experiment shows that simply rephrasing the whole sentence using BERT would lead to too diverse an output. Instead, a prior error span detection could be necessary for efficient GEC, and it is part of our future work.

\paragraph{Partial masking and fluency measures} Multi-masking or masking an informative part in the sentence will lead to loss of original information, and it will allow unwanted freedom in the predictions; see Table~\ref{tab: edit-error-analysis} for examples. Put in plain terms, multi-masking allows BERT to get too creative. Instead, we will investigate partial masking strategies~\cite{zhou-etal-2019-bert} which could alleviate this problem. 
Fluency is an important measure when employing an iterative approach \cite{napoles-sakaguchi-tetreault:2016:EMNLP2016}. We plan to explore fluency measures as part of our reranking mechanisms.
\bibliography{aaai}

\begin{thebibliography}{}

\bibitem[\protect\citeauthoryear{Bryant, Felice, and
  Briscoe}{2017}]{bryant-etal-2017-automatic}
Bryant, C.; Felice, M.; and Briscoe, T.
\newblock 2017.
\newblock Automatic annotation and evaluation of error types for grammatical
  error correction.
\newblock In {\em Proc. ACL}.

\bibitem[\protect\citeauthoryear{Choe \bgroup et al\mbox.\egroup
  }{2019}]{choe-etal-2019-neural}
Choe, Y.~J.; Ham, J.; Park, K.; and Yoon, Y.
\newblock 2019.
\newblock A neural grammatical error correction system built on better
  pre-training and sequential transfer learning.
\newblock In {\em Proc. BEA@ACL}.

\bibitem[\protect\citeauthoryear{Devlin \bgroup et al\mbox.\egroup
  }{2019}]{devlin-etal-2019-bert}
Devlin, J.; Chang, M.-W.; Lee, K.; and Toutanova, K.
\newblock 2019.
\newblock {BERT}: Pre-training of deep bidirectional transformers for language
  understanding.
\newblock In {\em Proc. NAACL-HLT}.

\bibitem[\protect\citeauthoryear{Ge, Wei, and
  Zhou}{2018}]{ge-etal-2018-fluency}
Ge, T.; Wei, F.; and Zhou, M.
\newblock 2018.
\newblock Fluency boost learning and inference for neural grammatical error
  correction.
\newblock In {\em Proc. ACL}.

\bibitem[\protect\citeauthoryear{Guu \bgroup et al\mbox.\egroup
  }{2017}]{Guu2017GeneratingSB}
Guu, K.; Hashimoto, T.~B.; Oren, Y.; and Liang, P.
\newblock 2017.
\newblock Generating sentences by editing prototypes.
\newblock {\em TACL} 6:437--450.

\bibitem[\protect\citeauthoryear{Napoles, Sakaguchi, and
  Tetreault}{2016}]{napoles-sakaguchi-tetreault:2016:EMNLP2016}
Napoles, C.; Sakaguchi, K.; and Tetreault, J.
\newblock 2016.
\newblock There's no comparison: Reference-less evaluation metrics in
  grammatical error correction.
\newblock In {\em Proc. EMNLP},  2109--2115.

\bibitem[\protect\citeauthoryear{Yannakoudakis, Briscoe, and
  Medlock}{2011}]{yannakoudakis-etal-2011-new}
Yannakoudakis, H.; Briscoe, T.; and Medlock, B.
\newblock 2011.
\newblock A new dataset and method for automatically grading {ESOL} texts.
\newblock In {\em Proc. ACL}.

\bibitem[\protect\citeauthoryear{Zhou \bgroup et al\mbox.\egroup
  }{2019}]{zhou-etal-2019-bert}
Zhou, W.; Ge, T.; Xu, K.; Wei, F.; and Zhou, M.
\newblock 2019.
\newblock {BERT}-based lexical substitution.
\newblock In {\em Proc. ACL}.

\end{thebibliography}
\bibliographystyle{aaai}
\end{document}